\pdfoutput=1

\documentclass[11pt]{article}

\usepackage{acl}

\usepackage{times}
\usepackage{latexsym}
\usepackage{amsfonts}
\usepackage{comment}
\usepackage{booktabs}
\usepackage{graphicx}
\usepackage{multirow}
\usepackage{longtable}
\usepackage{algorithm}
\usepackage{algorithmic}
\usepackage{amsmath}
\usepackage{colortbl}
\usepackage[T1]{fontenc}

\usepackage[utf8x]{inputenc}

\usepackage{microtype}
\usepackage{xcolor}
\usepackage{cleveref}
\crefname{section}{§}{§§}
\definecolor{c1}{HTML}{4e79a7}%
\definecolor{c2}{HTML}{f28e2b}%
\definecolor{c3}{HTML}{009E73}%
\definecolor{c4}{HTML}{56B4E9}%
\definecolor{c5}{HTML}{CC79A7}%
\definecolor{c6}{HTML}{E69F00}%
\definecolor{c7}{HTML}{844E4D}%
\definecolor{c8}{HTML}{2D512A}%

\definecolor{oorange}{HTML}{d95f02}
\definecolor{bblue}{HTML}{7570b3}
\definecolor{ggreen}{HTML}{1b9e77}
\definecolor{ppurple}{HTML}{e37fbb}
\definecolor{lgreen}{HTML}{9CD24A}
\definecolor{yyellow}{HTML}{FFD52D}
\definecolor{ggold}{HTML}{E1BC89}
\definecolor{ggray}{HTML}{AAAAAA}
\newcommand{\hl}[2]{\colorbox{#1!20}{{#2}}}
\usepackage[greek.polutoniko, english]{babel}
\DeclareUnicodeCharacter{0345}{\textsubscript{i}}

\usepackage{dirtytalk}
\usepackage{hyperref}

\title{Characterizing the Effects of Translation on Intertextuality using Multilingual Embedding Spaces}

\author{
    {\bf Hope McGovern}\textsuperscript{1} ~~
    {\bf Hale Sirin}\textsuperscript{2} ~~
    {\bf Tom Lippincott}\textsuperscript{2} ~~ \\
    \textsuperscript{1} Department of Computer Science \& Technology, University of Cambridge, U.K. \\
    \textsuperscript{2} Center for Digital Humanities, Johns Hopkins University, Baltimore, U.S.A. \\
    \small \textsuperscript{1} \texttt{hope.mcgovern@cl.cam.ac.uk}
    \hspace{3mm}
    \small \textsuperscript{2} \texttt{\{hsirin1, tom.lippincott\}@jhu.edu}
    \hspace{3mm} \\
}

\begin{document}
\maketitle
\begin{abstract}
Rhetorical devices are difficult to translate, but they are crucial to the translation of literary documents. We investigate the use of multilingual embedding spaces to characterize the preservation of intertextuality, one common rhetorical device, across human and machine translation. To do so, we use Biblical texts, which are both full of intertextual references and are highly translated works. We provide a metric to characterize intertextuality at the corpus level and provide a quantitative analysis of the preservation of this rhetorical device across extant human translations and machine-generated counterparts. We go on to provide qualitative analysis of cases wherein human translations over- or underemphasize the intertextuality present in the text, whereas machine translations provide a neutral baseline. This provides support for established scholarship proposing that human translators have a propensity to amplify certain literary characteristics of the original manuscripts.\footnote{All code and data available at \url{https://github.com/comp-int-hum/literary-translation}}

\end{abstract}

\section{Introduction}
Coined by the semiotician and literary critic Julia Kristeva in 1969, \emph{intertextuality} is a term that encompasses the ways in which one piece of text can refer to another \cite{kristev}. It can range from direct quotation to semantic resemblance, both within and between works, highlighting that ``no text is an island,'' and that a text can only be understood as part of a matrix of other texts, impacting both literary theory and translation theory that followed \cite{intertex_rev}. For example, intertextual allusions can be seen throughout James Joyce's retelling of Homer's \emph{Odyssey} in his 1922 novel, \emph{Ulysses}, realized through a broad range of linguistic and narrative correspondences \cite{currie}, such as the pairing between characters from each book: Molly/Penelope, Stephen/Telemachus and Leopold/Odysseus.

As earlier scholarship on computational detection of intertextuality points out, intertextual references have two main functions:
to express similarity between two passages, ``so that the latter can be interpreted in light of the former''; but also to highlight their differences, in that the earlier context they reference can be revised \cite{BammanCrane2008}. For example, in the film \textit{The Matrix} (1999) the white rabbit serves as an intertextual reference to \textit{Alice's Adventures in Wonderland} (1865) by Lewis Carroll. However, inverting the original context in which Alice was falling into a dream-land, Neo is now waking up from one. Intertextual references of this type set up a link akin to a two-way ``traffic''—inviting both similarities and differences \cite{hays1989echoes}. They are a prominent feature of Classical texts, notably the New Testament and its references to the Hebrew Bible \cite{BammanCrane2008}. \autoref{bliblical_intt} shows one such example.  Biblical intertextuality can range from explicit quotation to echoes of formulaic language, and many examples have since been cataloged \cite{hays1989echoes}. 


\begin{table*}

\begin{tabular}{p{0.49\linewidth} | p{0.45\linewidth}}
\hline
Exodus 14:21 &  Revelation 16:12\\

\textit{Hebrew Bible}  & \textit{New Testament} \\
\hline

 20 [...] and it was a cloud and darkness to them, but it gave light by night to these: so that the one came not near the other all the night. &  11 And blasphemed the God of heaven because of their pains and their sores, and repented not of their deeds. \\

\rowcolor[gray]{0.9} 21 \hl{ggreen}{And Moses stretched out his hand over the sea;} and the LORD caused the sea to go back by a strong east wind all that night, \hl{yyellow}{and made the sea dry land,} \hl{yyellow} {and the waters were divided.}& 12\hl{ggreen}{And the sixth angel poured his vial upon the} \hl{ggreen} {great river Euphrates;}\hl{yyellow}{and the water thereof} \hl{yyellow}{was dried up,}\hl{yyellow}{that the way of the kings of the} \hl{yyellow} {east might be prepared.}  \\
 22 And the children of Israel went into the midst of the sea upon the dry ground [...] & 13 And I saw three unclean spirits like frogs come out of the mouth of the dragon [...]

 \\

\hline
\end{tabular}
\centering

\caption{\small \textbf{Biblical intertextuality.} The highlighted middle verse shows the intertextual reference from the New Testament to the Hebrew Bible, establishing a connection between the drying up of the Euphrates River and Moses parting the Red Sea. Both are instances of divine intervention in the context of a body of water. However, intertextuality here not only establishes a semantic parallel between two events, but it also emphasizes the difference. The passage from Exodus is  moment of the divine judgment that leads to safety, whereas the drying up of the Euphrates is a preparation for the final judgment of the world. }
\label{bliblical_intt}
\end{table*}

 Detecting intertextual references contributes toward a contextualized understanding of the ``full semiotic density'' of a any given text \cite{broder} and therefore identifying intertextuality and the degree to which it is preserved in translation is crucial for the interpretation and appreciation of literary and historical texts. Due to its significance, computational methods for identifying intertextuality have become an expanding field of research, and it is closely connected to other NLP tasks that are grouped under narrative reasoning and comprehension \cite{Sang2022ASO, pipertheory}.
 Significant attention has been devoted to identifying text reuse (implicit intertextuality) in Biblical text \cite{LeeComputationalModelText2007, MoritzNonLiteralTextReuse2016a}, classical Latin poetry \cite{burns2021profiling, BammanCrane2008}, Latin prose \cite{dexterQuantitativeCriticismLiterary2017}, and Romantic poetry \cite{ForstallS19}. Several of these works, \citet{burns2021profiling} in particular, highlight that neural embeddings can be used effectively to capture intertextuality. However, not much attention has been paid to the \textit{effects of translation} on intertextual references.  





In this work, we look at translation effects on intertextuality in the Bible through neural embedding spaces. While the Bible is often treated as one text, it is in fact a library of texts written by an estimated 60 different authors over the course of 4,000 years, and therefore offers a unique test bed for the detection of intertextuality and the effects of translation. This is especially true given the multilingual nature of the intertextuality between the New Testament and the Hebrew Bible in their original Greek and Hebrew. 

Our main contributions are as follows:

\begin{enumerate}
    \item We show that multilingual embedding spaces may be effectively used to characterize intertextuality in original documents as well as their translations.
    \item We provide a new method for characterizing intertextuality within and across translations.
    \item We conduct a comparative study of human- and machine-generated translations of the same corpus into different languages of varying resource levels.
    \item We contribute to Classical and Biblical scholarship that qualitatively explores whether human translations have, purposefully or not, amplified intertextuality between the old and new testaments for the sake of continuity\footnote{For instance, Erich Auerbach underlines that Paul’s historical mission among the Gentiles needed to separate Christianity from Judaism by conveying the idea that ``the old Law is suspended and replaced'' through references that both alluded to \textit{and} recontextualized the Hebrew Bible \cite{figura} \cite{Sirin_diss}.}.

\end{enumerate}

\section{Characterizing intertextuality between Corpora}
\label{sec:formalism_intertext}

Our intertextuality measure is simply the cosine similarity of a pair of verse embeddings from a multilingual embedding model. For a given set of ground-truth references, we can also compute \emph{baseline} similarities by randomly swapping one of the verses with another from the same chapter\footnote{Maintaining the same chapter ensures that false pairs likely remain upon the same topic, as opposed to choosing a random verse from anywhere in the Bible.}. The ratio of the average intertextuality similarity to the average baseline similarity 
can be used to compare the degree of intertextuality across different sets of translations. 

Intuitively, a ratio much larger than one (1) indicates strong intertextuality, whereas anything less than one indicates that supposedly intertextual verses are not more similar than random pairings. When comparing changes in intertextuality ratio across translation, we compute the 95\% confidence interval via bootstrapping. Specifically, we resample the original data with replacement 10,000 times, recalculating the ratio for each resample. 

Note that this method relies upon having access to ground-truth references — or suspected references — and would likely be too crude a method to discover novel instances of intertextuality without extensive threshold tuning. Instead, we use this measure to ascertain the \emph{degree} of intertextuality within a set of texts known to be intertextual. We can then use this measurement to characterize changes in intertextuality across the same set of texts in translation.

We compute intertextuality ratios for all original, human, and machine-translation texts, distinguishing the sets of references that are internal to a testament (\emph{within}) and that cross between them (\emph{across}). This distinction allows us to consider whether Christian writing is particularly referential to the Jewish Testament, or if it became so through the effects of translation. Christian theologians throughout history have often underscored the continuity of the Christian and Jewish testaments \cite{continuity}, and human translators may have sought to emphasize this continuity in their translations. The full tables of these ratios can be found in \autoref{table:intt_scores}.

\section{Method}
\begin{table}[t]
\centering
\small
\begin{tabular}{lll}
\hline
Language & Family & Bitext pairs \\
\hline
English & West Germanic & \( > 10M \) \\
Finnish & Uralic & \( > 1M \) \\
Turkish & Turkic & \( > 100K \) \\
Swedish & North Germanic & \( > 10K \) \\
Marathi & Indo-Aryan & Small \\
\hline
\end{tabular}
\centering
\caption{\small \textbf{Languages by family.} Summary of languages used in this study: each has a full, aligned human translation of both the Jewish and Christian texts.
The sizes are reported from \citet{mbart50} training data and reflect the variety of resource-levels.}
\label{table:languages}
\end{table}

\subsection{Data}
We use three primary sources for our textual analysis: the Translator’s Amalgamated Hebrew Old Testament (TAHOT)
and Greek New Testament (TAGNT)\footnote{\url{www.STEPBible.org}}, as well as a digitized copy of the Septuagint (LXX)\footnote{\small\url{https://sourceforge.net/projects/zefania-sharp/files/Bibles/GRC}}.
The TAHOT is based on the Leningrad Codex, the oldest complete extant version of the Hebrew Old Testament. The TAGNT consolidates the Greek New Testament text from multiple early extant editions, and these are both compiled by Bible scholars at Tyndale House in Cambridge, UK, and released as part of the STEP Bible project\footnote{We release a formatted version of STEP Bible's data on the Huggingface Hub. DOIs: \href{https://huggingface.co/datasets/hmcgovern/original-language-bibles-hebrew}{10.57967/hf/4174}, \href{https://huggingface.co/datasets/hmcgovern/original-language-bibles-greek}{10.57967/hf/4184}.}. The Septuagint is the earliest Greek translation of the Hebrew Old Testament, completed by Jewish scribes in the few centuries preceding the events of the New Testament.\footnote{
The complex history of Biblical scribal tradition means that almost all modern English translations use a versification system which at many points differs from the versification in the original Hebrew (cf. Genesis 31:55 in English translations is considered Genesis 32:1 in the Leningrad Codex). For consistency, we align all documents to use the English versification system across all experiments.}

For modern human translations, we use the Johns Hopkins University Bible Corpus \cite{McCarthyJohnsHopkinsUniversity2020} for the five languages in \autoref{table:languages}, each of which include both testaments.

To independently evaluate our method, we use a benchmark corpus for intertextuality provided by \citet{burns2021profiling} detailing intertextual references in Classical Latin literature. Specifically, it contains 945 references curated by subject matter experts connecting Valerius Flaccus' \textit{Argonautica I} to earlier and contemporary Roman authors.

\begin{table}[]
\footnotesize
\small
\centering
\begin{tabular}{@{}clll@{}}
\toprule
\multicolumn{1}{c}{\multirow{2}{*}{}} & \multicolumn{3}{c}{Source Manuscript} \\
\multicolumn{1}{c}{Target}                                 & Hebrew OT   &       Greek OT   &     Greek NT   \\ \midrule
English                                              &  \textbf{69.5} &   \textbf{61.2} &  \textbf{72.6}      \\
Finnish                                              &   47.6    &  43.9  &  48.8  \\
Turkish                                              &   \textit{66.7}          &   \textit{65.4}         &    \textit{68.2}        \\
Swedish                                              &   54.0          &   53.7         &     56.0       \\
Marathi                                              &   27.6          &  26.5          &   29.8 \\
\bottomrule
\end{tabular}
\caption{\small \textbf{COMET scores}. Top-scoring translation for each source manuscript is in bold text. Second top-scoring translation is in italics.}
\label{tab:comet}
\end{table}
\subsection{Translation}
\label{sec:translation}
To compare the effects of human and machine translation,
we employ Cohere's multilingual model Aya23\footnote{We use the model version with 8B parameters.} \cite{aryabumi2024aya} to translate all of the original manuscripts into the five languages of varying resource levels from \autoref{table:languages}. Aya23 is chosen for this task as it has been shown to outperform other multilingual models of similar, and sometimes larger, sizes for machine translation \cite{aryabumi2024aya}, but is small enough to be practical for academic research settings with limited compute power. Full pre-processing, prompting, and post-processing details may be found in \cref{sec:appendix}.
We report translation quality scores using the COMET metric \cite{rei2020comet} in \autoref{tab:comet}, providing references (human-translated text in the target language), predictions (machine-generated text in the target language), and sources (original text in the original language). 

\subsection{Gold standard for intertextuality}
For ground-truth information about which passages are truly interlinked, we use a dataset of Bible cross-references \cite{references}. According to the dataset's documentation, the initial data was seeded largely from the Treasury of Scripture Knowledge \cite{torrey1982treasury}, an authoritative compilation of cross-references from prominent Biblical scholars over many centuries, which was then cleaned to remove duplicates and concatenate separate entries for adjacent references. Finally, the references were opened to crowd-sourcing annotation for voting on relevant connections.

We limit consideration to verse-to-verse links that connect passages from different books and can be resolved in all manuscripts. We disregard ordering by 
summing the votes for both directions between a pair of verses, and use a vote threshold of \( 50 \) to consider a reference valid.\footnote{We independently verify that $96.0\%$ of the cross-references in our dataset with at least 50 votes are attested in an online version of the Treasury of Scripture Knowledge \url{https://www.tsk-online.com/}.}
This produces a total of \( 2183 \) references: \( 548 \) are entirely within the Jewish testament, \( 961 \) within the Christian, and \( 674 \) that span them. 
We differentiate these two cases with the qualifiers \emph{within}, meaning within the same testament, and \emph{across}, meaning across the two testaments.

\begin{table}[H]
\tiny
\centering
\begin{tabular}{lllll}
\toprule
& & \multicolumn{2}{c}{Within} & Across \\
&& Jewish (OT) & Christian (NT) & \\
\midrule
\multirow{2}{*}{\rotatebox[origin=c]{90}{\tiny Orig. }}&Ancient Hebrew & 0.98$\pm$ 0.14& --&  --\\
& Ancient Greek & 1.27$\pm$ 0.21 & 1.30$\pm$0.19 & 1.31$\pm$ 0.20 \\ 

\midrule
\multirow{5}{*}{\rotatebox[origin=c]{90}{\tiny Human}}& English & 1.66$\pm$ 0.21 & 1.70$\pm$ 0.30 & 1.69$\pm$ 0.27\\
&Finnish & 1.42$\pm$ 0.41 & 1.36$\pm$ 0.53 & 1.48$\pm$ 0.22 \\
&Turkish & 1.50$\pm$ 0.18 & 1.43$\pm$ 0.29 &  1.51$\pm$ 0.12 \\
&Swedish & 1.33$\pm$ 0.15 &  1.39$\pm$ 0.12 & 1.37$\pm$ 0.09 \\
&Marathi & 1.35$\pm$ 0.12 & 1.42$\pm$ 0.12 & 1.44$\pm$ 0.10\\ \midrule
\multirow{5}{*}{\rotatebox[origin=c]{90}{\tiny NMT }}&English & 1.32$\pm$ 0.20 & 1.50$\pm$ 0.17 & 1.48$\pm$ 0.20 \\
&Finnish & 1.24$\pm$ 0.22 & 1.28$\pm$ 0.18 & 1.26$\pm$ 0.11 \\
&Turkish & 1.60$\pm$ 0.15 &  1.71 $\pm$ 0.12 & 1.52$\pm$ 0.32 \\
&Swedish & 1.31$\pm$ 0.22 & 1.29$\pm$ 0.31 & 1.36$\pm$ 0.30 \\
&Marathi & 1.02 $\pm$ 0.10 & 1.22$\pm$ 0.25 &  1.30$\pm$ 0.18\\

\bottomrule
\end{tabular}

\caption{\small \textbf{Intertextuality ratios for source manuscripts and their human translations.} Ratios within, and where possible between, testaments, for the Septuagint and TAGNT (Greek), TAHOT (Hebrew), and five human translations with 95\% CI.}
\label{table:intt_scores}
\end{table}
\section{Experiments}
\textbf{Benchmark Corpus}:  First, we evaluate our method on a benchmark corpus for intertextuality between Valerius Flaccus' \textit{Argonautica I} to earlier and contemporary Roman authors. We calculate an intertextuality ratio of 1.55, 95\% CI [1.53,1.56], indicating that our method succeeds at characterizing known intertextuality at the corpus level. 

\textbf{Translation Quality}:  \autoref{tab:comet} shows translation scores from the Hebrew Old Testament, Greek Old Testament, and Greek New Testament into five target languages. English and Turkish consistently achieve the highest scores across all manuscripts, with English translations ranging from 61.2 to 72.6, and Turkish from 65.4 to 68.2, suggesting strong translation quality for these language pairs. In contrast, translations into Marathi show the lowest scores, ranging from 26.5 to 29.8, likely due to the complexity of translating between less common language pairs. These results establish a valuable benchmark for evaluating translation quality for underrepresented languages in historical texts.
\begin{table}
\footnotesize
\begin{tabular}{lp{.4\textwidth}}

\toprule
a & \textgreek{ὅτι ἵλεως ἔσομαι ταῖς ἀδικίαις αὐτῶν, καὶ τῶν {\color{red}ἁμαρτιῶν} αὐτῶν οὐ μὴ μνησθῶ ἔτι.} \\
\cline{2-2}
b & \textgreek{ἐγώ εἰμι ἐγώ εἰμι ὁ ἐξαλείφων τὰς ἀνομίας σου καὶ οὐ μὴ μνησθήσομαι.} \\
\midrule
a & For I will be merciful to their unrighteousness, and their {\color{red}sins} and their iniquities will I remember no more. \\
\cline{2-2}
b & I, even I, am he that blotteth out thy transgressions for mine own sake, and will not remember thy {\color{red}sins}. \\
\cline{2-2}
a' & For I will beware of their iniquity, and their {\color{red}sinner's} iniquity; for I will not abhor them: \\
\cline{2-2}
b' & I am the last of thy iniquitous acts, and I hate not myself. \\
\bottomrule
\end{tabular}
\caption{\small \textbf{Overemphasized Intertextuality by Human Translation.} The intertextuality from Hebrews 8:12 to Isaiah 43:25 is amplified by the human translator's decision to render different words as "sin". The machine translation abstains from this and restores the original distance, but loses coherence.}
\label{table:qual}
\end{table}

\section{Analyzing Intertextuality in Translation} 

\autoref{table:intt_scores} shows that there is a higher degree of intertextuality \textit{across} the New Testament and the Hebrew Bible compared to intertextual references \textit{within} each book. 

The degree to which intertextuality is preserved is highest for the English translation and lowest for Marathi. Human translations consistently show higher levels of intertextuality. 


As suggested by \citet{mcgovern-etal-2024-typescene}, we indeed see that human translations over or underemphasize the intertextuality present in the text, whereas machine translations provide a neutral baseline, based on these results.
We can look closer at the translation effects by sorting intertextual pairs according to the absolute shift in similarity. \autoref{table:qual} shows the original Greek, the human English translation, and the unconstrained machine translation for one such pair, between the Epistle to the Hebrews and the Book of Isaiah. The pair of verses has strong similarity in the original Greek (0.332), but this is nearly doubled by the human English translation (0.656). The highlighted Greek word, \emph{hamartion}, typically translated as \emph{sin}, occurs in Hebrews but not Isaiah, yet the latter's human translation makes a point of using the term. Surface-level lexical decisions like this, and presumably many less direct choices, lead to uncalibrated translations that reinforce the received interpretation.



\section{Future work}
We plan to address the persistent issue of misalignment in parallel Bible corpora. Even in scholarly editions of digitized texts, misalignment is persistent. However, by applying the alignment methodology proposed by \cite{Crane_Align}, we could unify alignment for research purposes. Finally, we leave to future work exploring larger narrative contexts by examining narrative episodes instead of verse-level intertextuality.


\section*{Limitations}

In this work, we generate machine translations working from the oldest extant manuscripts of the Biblical texts. However, most translations present in the JHUBC were not translated directly from ancient manuscripts but instead work from English translations, which themselves were often translations of the Greek texts. So direct comparisons of the human translations and machine translations in this work should be treated with caution.

\section*{Acknowledgments}
We would like to thank Matt Post for very helpful discussions about machine translation evaluation, as well as scholars at Tyndale House, especially Ellie Wiener and Caleb J. Howard, for useful discussions about Biblical Hebrew. Thanks to Andrew Caines for his helpful comments on previous drafts.
Hope McGovern's work is supported by the Woolf Institute for Interfaith Relations and the Cambridge Trust.

\bibliography{literary_translation}
\bibliographystyle{acl_natbib}
\onecolumn

\appendix
\label{sec:appendix}
\section{Additional Implementation Details}
\textbf{Preprocessing}
We use CohereForAI's Aya-23 8B model to generate all machine translations. We do not remove any accents or diacritics as preprocessing. 

\textbf{Prompting}
We use few-shot prompting to obtain our translations.
an example prompt can be seen below:

``Translate the following Ancient Greek phrases into English:

1. Ancient Greek: ``\textgreek{εἰ  δέ  τις  ἐποικοδομεῖ  ἐπὶ  τὸν  θεμέλιον  τοῦτον  χρυσόν,  ἄργυρον, λίθους  τιμίους,  ξύλα,  χόρτον,  κα
     λάμην,}''
     
English: ``Now if any man build upon this foundation gold, silver, precious stones, wood, hay, stubble;''

2. Ancient Greek: ``\textgreek{καὶ  οὐθὲν  διέκρινεν  μεταξὺ  ἡμῶν  τε  καὶ  αὐτῶν  τῇ  πίστει καθαρίσας  τὰς  καρδίας  αὐτῶν.}''

English: ``And put no difference between us and them, purifying their hearts by faith.''

3. Ancient Greek: ``\textgreek{εἰ  δὲ  Χριστὸς  οὐκ  ἐγήγερται,  κενὸν  ἄρα  καὶ  τὸ  κήρυγμα ἡμῶν,  κενὴ  δὲ  καὶ  ἡ  πίστις  ὑμῶν.}''

English: ``And if Christ be not risen, then is our preaching vain, and your faith is also vain.''

4. Ancient Greek: ``\textgreek{καὶ  ἐν  τούτῳ  γνωσόμεθα  ὅτι  ἐκ  τῆς  ἀληθείας  ἐσμὲν  καὶ ἔμπροσθεν  αὐτοῦ  πείσομεν  τὴν  καρδίαν  ἡμῶν}''

English: ``And hereby we know that we are of the truth, and shall assure our hearts before him.''

Now, translate this Ancient Greek phrase:

5. Ancient Greek: ``INPUT\_TEXT''

English:''

For the prompt, we draw four (4) examples of translations from the source texts. Ideally, these translations would be drawn from other parallel sources, but for most of the translation pairs (e.g. Ancient Hebrew $\rightarrow$ Marathi), the Biblical texts are the only parallel data available.

\textbf{Generation} At inference time, we use a maximum output length of 100 new tokens. We use the default BPE tokenizer with all of the default settings. 

\textbf{Post-Processing}
We find that we need to post-process the outputs: we grab what is in the first set of quotation marks after our prompt and exclude the rest. We find this is necessary to prevent nonsensical continued generations.

\texttt{N.B.} Models were access through the Huggingface Transformers library \cite{wolf2020huggingfaces}.


\end{document}